\documentclass[letterpaper, 10 pt, conference]{ieeeconf}
\IEEEoverridecommandlockouts                              

\usepackage{tikz}
\usetikzlibrary{shapes}
\usepackage{physics}
\usepackage{amsmath, amssymb}
\usepackage{bbm}
\usepackage{algorithmic}
\usepackage{algorithm}
\usepackage{subfig}
\usepackage[colorinlistoftodos, textwidth=2cm,textsize=tiny]{todonotes}
\usepackage{siunitx}
\usepackage{flushend}

\title{\LARGE \bf
Gaussian-Dirichlet Random Fields for Inference over High Dimensional Categorical Observations
}

\author{John E. San Soucie$^{1}$ Heidi M. Sosik$^{2}$ and Yogesh Girdhar$^{3}$
\thanks{*This work was supported by a grant from the Simons Foundation (561126, HMS) and NSF-NRI Award Number 1734400}
\thanks{$^{1}$J. San Soucie is with the Mechanical Engineering department at the Massachusetts Institute of Technology and the Applied Ocean Physics and Engineering Department at the Woods Hole Oceanographic Institution
        {\tt\small johnes@mit.edu}}%
 \thanks{$^{2}$H. Sosik is with the Biology Department at the Woods Hole Oceanographic Institution
        {\tt\small hsosik@whoi.edu}}%
 \thanks{$^{3}$Y. Girdhar is with the Applied Ocean Physics and Engineering Department at the Woods Hole Oceanographic Institution
        {\tt\small yogi@whoi.edu}}%
}

\def\doi{DOI: \href{https://doi.org/XX.YYYYY/X.YYYYY}{XX.YYYYY/X.YYYYY}}
\newcommand\copyrighttext{%
  \footnotesize \textcopyright2020 IEEE. Personal use of this material is permitted.
  Permission from IEEE must be obtained for all other uses, in any current or future
  media, including reprinting/republishing this material for advertising or promotional
  purposes, creating new collective works, for resale or redistribution to servers or
  lists, or reuse of any copyrighted component of this work in other works.
  }
\newcommand\copyrightnotice{%
\begin{tikzpicture}[remember picture,overlay]
\node[anchor=south,yshift=10pt] at (current page.south) {\fbox{\parbox{\dimexpr\textwidth-\fboxsep-\fboxrule\relax}{\copyrighttext}}};
\end{tikzpicture}%
}

\begin{document}

\maketitle
\thispagestyle{empty}
\pagestyle{empty}

\begin{abstract}
We propose a generative model for the spatio-temporal distribution of high dimensional categorical observations. These are commonly produced by robots equipped with an imaging sensor such as a camera, paired with an image classifier, potentially producing observations over thousands of categories. The proposed approach combines the use of Dirichlet distributions to model sparse co-occurrence relations between the observed categories using a latent variable, and Gaussian processes to model the latent variable's spatio-temporal distribution. Experiments in this paper show that the resulting model is able to efficiently and accurately approximate the temporal distribution of high dimensional categorical measurements such as taxonomic observations of microscopic organisms in the ocean, even in unobserved (held out) locations, far from other samples. This work's primary motivation is to enable deployment of informative path planning techniques over high dimensional categorical fields, which until now have been limited to scalar or low dimensional vector observations.
\end{abstract}

\section{Introduction}
\copyrightnotice
The field of autonomous robotics has drastically increased the ability of scientists to explore remote environments \cite{Bellingham2007}. Sites of research interest in outer space and the deep ocean are characterized by being extremely inhospitable to human life and difficult to reach from the Earth's surface via radio-frequency communications. But by making higher-level decisions on-board, autonomous agents reduce the need for real-time communications with a human operator. The effectiveness of any autonomous agent, in terms of valuable information acquired per unit resource, depends largely on the plan it is programmed to follow. But remote environments are often under-explored, and thus the optimal plan is not known before the mission starts. Therefore, autonomous agents require plans that adapt to the environments they encounter, or informative path planning (IPP).

Robots performing adaptive sampling or IPP missions typically operate over measurements of interest that are low dimensional. For example, in the context of underwater robotic missions, IPP has been used to adaptively sample physical \cite{Cruz2010AdaptiveVehicles,Zhang2012UsingFront,Zhang2001Spectral-featureVehicle}, chemical \cite{Saigol2010a,Farrell2003ChemicalAUV,CamilliIntegratingSystems,JakubaExplorationVehicle}, biological \cite{Godin2011PhytoplanktonVehicle,Fossum2019}, and general visual signals \cite{Jamieson2020} in the ocean (for more details, see \cite{Hwang2019AUVReview}). IPP techniques require a spatial model for the information that is being sampled \cite{Binney2012BranchPlanning}. Gaussian processes (GPs) \cite{Williams,Rasmussen:2005:GPM:1162254}, which are capable of modeling black-box scalar or vector functions, are a common choice of probabilistic observation model for IPP.  GP models enable IPP algorithms to produce smooth estimates over future observations, along with uncertainties, which can then be used to predict expected trajectory reward and to balance the exploration-versus-exploitation trade-off using Bayesian Optimisation (BO) \cite{Marchant2014}.

\begin{figure}
    \centering
    \includegraphics[width=0.9\columnwidth]{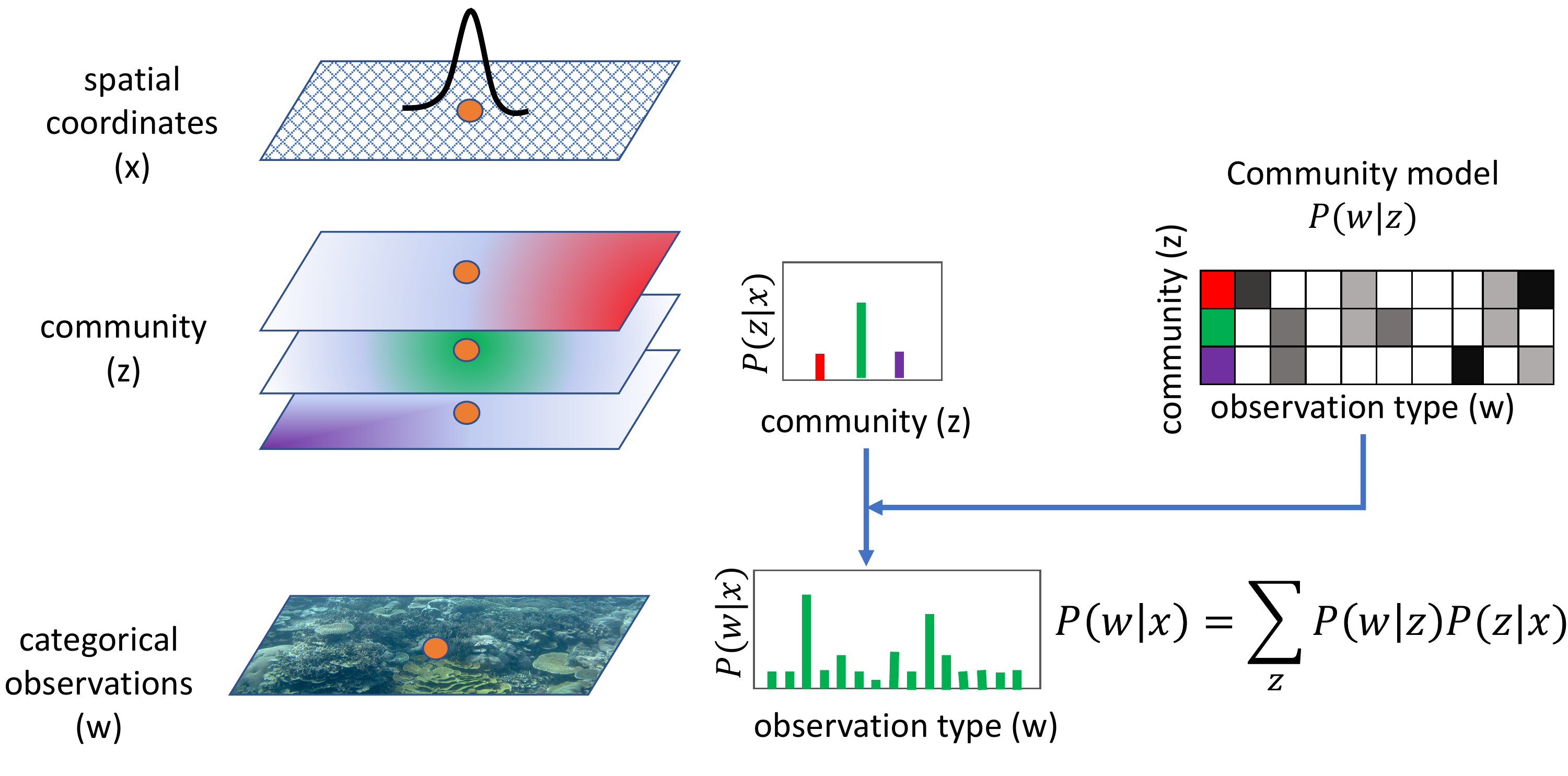}
    \caption{Overview of the Gaussian-Dirichlet Random Field model. Categorical observations, such as observations of phytoplankton taxa, are factored into the product of a community model and spatiotemporal distributions for each community. The community model, which is the distribution of taxa in each community, is modeled with a Dirichlet prior; and the spatial distribution of each community is modeled using a Gaussian process.}
    \label{fig:gdrf_overview}
\end{figure}

Recent advances in machine learning and artificial intelligence have enabled development of new types of sensing systems that effectively measure high-dimensional categorical data. These systems combine traditional sensors (cameras, probes, etc.) with deep neural networks to classify scalar, vector, and image-based measurements in real time \cite{DBLP:journals/corr/RedmonDGF15, Szegedy2017}. Examples of such systems include classifiers trained using datasets \cite{Orenstein2015} produced by the Imaging FlowCytobot (IFCB) \cite{Sosik2007}, that can produce taxonomic observations of phytoplankton with over 100 categories, and classifiers trained on the ImageNet dataset \cite{Russakovsky2015}, which can produce categorical observations over 10,000 categories. IPP techniques based on GP models do not directly generalize to such high-dimensional categorical observations
as common embedding methods (such as one-hot encoding\cite{Garrido-Merchan2016a}) embed a $K$-categorical variable in $K$ dimensions, leading to exponential space complexity. Hierarchical Bayesian models designed for spatio-temporal data, such as HDP-ROST \cite{Girdhar2016}, can efficiently model distributions of high dimensional categorical observations. But these models do not provide smoothly varying probability distribution estimates at arbitrary locations in space and time, which are needed for BO based IPP. Thus, there is a need for a spatio-temporal model of the distribution of high-dimensional categorical data compatible with IPP.



In this paper, we introduce the Gaussian-Dirichlet Random Field (GDRF), a hierarchical generative topic model for the spatial distribution of categorical observations. As shown in fig. \ref{fig:gdrf_overview}, GDRFs factor the probability distribution over observation categories $w$ into conditional distributions for $w$ given a latent topic $z$, and conditional distributions of the latent topics given the observation's location $x$:
\begin{equation}
    P(w|x) = \sum_z P(w|z) P(z|x)
\label{P(w|x)}
\end{equation}
Since $w$ and $z$ are both categorical random variables, \eqref{P(w|x)} remains unusable for IPP. Therefore, we further factor $P(z|x)$ by introducing a latent Gaussian random field $\mu_i$ for each topic, normalized to a distribution via a link function $f_i$: 
\begin{equation}
P(z_i|x) = f_i(\mu_1(x), \dots, \mu_K(x))
\label{P(z_i|x)}
\end{equation}
To learn the word-topic model and the latent $\mu_i$, we combine Gibbs sampling with variational inference. An immediate consequence of our choice of factorization is that the topics in a GDRF model are scientifically meaningful in that they capture the latent spatiotemporal structures relating different observation categories.

\section{Related Work}

Latent Dirichlet Allocation (LDA), introduced in \cite{blei2003} is a topic model \cite{Blei2012ProbabilisticModels} originally designed for text documents. LDA models the relationships between words and documents using a set of latent topics that are linked to both the words and documents via Dirichlet distributions. Formally, given a corpus of $M$ documents $\{d_1, \dots, d_M\}$, each with $N_i$ words out of a vocabulary of size $W$, we can generate a topic model for $K$ topics as follows:
\begin{align}
\Theta_{d_i}&\sim \text{Dirichlet}(\alpha) \nonumber\\
\Phi_{z_k} &\sim \text{Dirichlet}(\beta) \nonumber\\
z_{j, d_i} &\sim \text{Cat}(\Theta_{d_i}) \nonumber \\
w_{j, d_i} &\sim \text{Cat}(\Phi_{z_{j, d_i}}) \label{LDA}
\end{align}
\begin{figure}[h!]
    \centering
    \begin{tikzpicture}

\node (theta) at (0,0) [circle, draw] {$\Theta$};
\node (z) at (2,0) [circle, draw] {$z$};
\node (w) at (4,0) [circle, draw, fill=lightgray] {$w$};
\node (alpha) at (0,2.4) [regular polygon,regular polygon sides=4, draw] {$\alpha$};
\node (psis) at (4,2.4) [circle, draw] {$\Phi$};
\node (b) at (2.5,2.4) [regular polygon,regular polygon sides=4, draw] {$\beta$};

\draw[->] (alpha.south) -- (theta.north);
\draw[->] (b.east) -- (psis.west);

\draw[->] (theta.east) -- (z.west);
\draw[->] (z.east) -- (w.west);
\draw[->] (psis.south) -- (w.north);

\draw (-1,-1) rectangle (6,1.5);
\draw (1,-.5) rectangle (5,1.2);
\draw (1,-.5) rectangle (5,1.2);
\draw (3.2,1.8) rectangle (4.8,3.0);

\node at (5.5,-.5) {$M$};
\node at (4.8,-.3) {$N_i$};
\node at (4.5,2.7) {$K$};
\end{tikzpicture}
    \caption{The graphical model for LDA}
    \label{fig:graphical_LDA}
\end{figure}
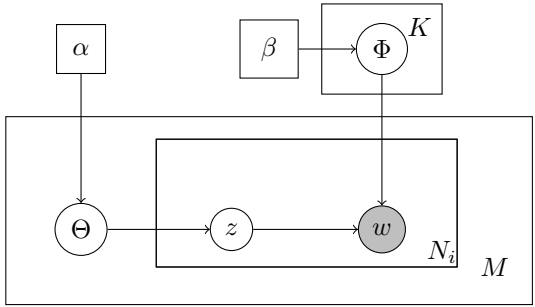\\
The graphical model for LDA is shown in Fig. \ref{fig:graphical_LDA}.
To perform inference, we can do Gibbs sampling for the topic $z_i$ assigned to a particular word $w_i$ in document $d_i$ \cite{griffiths2004}:
\begin{equation}
	P(z_i = j | \vb{z}_{-i}, \vb{w}) \propto \frac{n_{-i,j}^{w_i}+\beta}{n_{-i,j}^{\cdot}+W\beta} \frac{n_{-i,j}^{d_i}+\alpha}{n_{-i,\cdot}^{d_i}+K\alpha}
\label{LDA_gibbs}
\end{equation}
where $n_{-i,j}^{w_i}+1$ is the number of words with label $w_i$ assigned to topic $j$, $n_{-i,j}^{\cdot}+1$ is the total number of words assigned to topic $j$, $n_{-i,j}^{d_i}+1$ is the number of words in documents $d_i$ assigned topic $j$, and $n_{-i,\cdot}^{d_i}+1$ is the number of words in document $d_i$.
LDA has been applied to model natural scenes \cite{Fei-Fei2005ACategories} and human actions \cite{Niebles2008UnsupervisedWords}, in addition to text corpora such as electronic health records \cite{Hripcsak2013}, Twitter posts \cite{Vosoughi2018TheOnline}, and historical documents \cite{Newman2006ProbabilisticNewspaper}. Spatial LDA \cite{NIPS2007_3278} (SLDA) extends the LDA model to account for spatial structures in image data by adding priors over the location of a spatial word.

The Real-time Online Spatiotemporal Topic model (ROST), introduced in \cite{Girdhar2016}, models spatiotemporally distributed categorical data by discretizing an N-dimensional world and treating cells in that world as documents. Letting $G(d_i)$ be the spatiotemporal neighborhood of cell $d_i$, a hypothetical generative model for ROST simply replaces the distribution of topics for a document with the distribution of topics in the neighborhoodof a cell:
\begin{align}
\Theta_{G(d_i)}&\sim \text{Dirichlet}\qty(\alpha)
 \label{ROST}
\end{align}
The Gibbs sampling step for inference similarly replaces the count of topics in a document $n_{-i,j}^{d_i}$ with the count of topics in the neighborhood of a $n_{-i,j}^{G(d_i)}$.

ROST has been used for robots displaying unsupervised curious behavior \cite{Girdhar2014b, Girdhar2016}, multi-robot topic modeling \cite{Doherty2018ApproximateCharacterization} and phytoplankton ecological modeling \cite{Kalmbach2017b}.

Formally, a Gaussian process \cite{Rasmussen:2005:GPM:1162254} is a set of random variables $\{Z\qty(x_i)\}$ defined on some possibly infinite indexing set $X = \{x_i\}$ such that for any finite subset $Y \subseteq \{Z\qty(x_i)\}$, $Y \sim \mathcal{N}\qty(\vb*{\mu}, \vb*{\Sigma})$. The function $\Sigma: X \times X \to \mathbb{R}_{\ge 0}$ is called the \textit{kernel} or \textit{covariance} function of the GP, and specifies the structure of relationships between different points. Many common kernel functions on $\mathbb{R}^D$ are stationary ($\Sigma = \Sigma\qty(x-x')$) and isotropic  ($\Sigma = \Sigma\qty(|x-x'|)$, including the Mat\'ern kernel function with $\nu=3/2$:
\begin{equation}
k_{3/2}(r) = \sigma\qty(1+\frac{\sqrt{3}r}{\ell})\exp{-\frac{\sqrt{3}r}{l}}
\label{matern}
\end{equation}
In \eqref{matern}, $\ell$ represents the \textit{length scale} of the Gaussian process, and $\sigma$ represents a scale parameter for the kernel.

The geostatistics literature has used Kriging and similar spatial optimal linear predictation tools since the 1950s \cite{Matheron1963,Noel1990}. In Kriging, the value of a scalar field is modeled as a random field $Z(x)$. The value of $Z(x_0)$, an unobserved location, is estimated from a weighted sum of $n$ sampled locations:
\begin{equation}
\hat{Z}(x_0) = \sum_{i=1}^{n}\lambda_i Z(x_i)
\label{kriging}
\end{equation}
The weights that provide the minimum variance unbiased estimator $\hat{Z}(x_0)$ can be calculated given only the sampled values and the covariance function for the random field. Standard Kriging is a form of Gaussian Process regression \cite{Williams}. For categorical random fields, geostatisticians use variations on indicator Kriging \cite{Journel1983,Olea1999}. Methods from the Kriging family have been used for adaptive sampling \cite{Fentanes2018} and marine visual data analysis \cite{Jerosch2006}.

There are many examples in the literature of IPP algorithms which utilize GPs as a model for observed scalar fields in marine environments. Binney, Krause, and Sukhatme \cite{Binney2013OptimizingPhenomena} demonstrate a graph-based submodular optimization IPP algorithm with an objective function that can handle temporal nonstationarity and along-path sample collection (as opposed to sample collection at waypoints). Das et al. \cite{Das2013HierarchicalPhenomena} use GPs to model both spatial scalar fields and the relationship between the scalar field's variables and organism abundance. They tested their model with two adaptive sampling strategies. Suryan and Tokekar \cite{Suryan2018} develop a fast GP regression informative path planning algorithm over spatial fields. Berget et al. \cite{Berget2018AdaptiveModel}, Fossum and Eidvsik et al. \cite{Fossum2018Information-drivenOcean}, and Fossum and Fragoso et al. \cite{Fossum2019} implement GP regression on AUVs and demonstrate the viability of simple IPP algorithms in real-world scenarios. Flaspohler et al. \cite{Flaspohler2019} introduce a plume-finding algorithm, which locates maxima of phenomena modeled by GPs. In all of these works, GPs model scalar fields. 
\section{Gaussian-Dirichlet Random Fields}
We begin with several preliminary definitions. A GDRF is defined on an \textit{indexing set} $\vb{X} = \{x_1, x_2, \dots\}$, representing points in the world on which the model is defined. For example, a GDRF on a two-dimensional $A\times B$ grid has as its indexing set $\vb{X} = \{1, 2, \dots, A\} \times \{1, 2, \dots, B\}$. We will generally refer to the indexing set itself as the \textit{world}, and call members of the world \textit{locations}.

\textit{Words} $w_i$ and \textit{topics} $z_i$ are $W$- and $K$-categorical variables, respectively. The \textit{mean latent log probabilities} (MLLPs) $\mu_j$ are Gaussian random fields defined on the world $X$. MLLPs are transformed to a probability distribution via a \textit{link function} $f_j: \mathbb{R}^K \to \qty[0, 1]$, where $\sum_jf_j(\mu_1, \dots, \mu_K) = 1$ . For this paper, we exclusively use the softmax link function $f_j(\mu_1, \dots, \mu_K) = \exp(\mu_j)/\sum_k\exp(\mu_k)$. Finally, the generative model contains several hyperparameters: $\beta$ is the Dirichlet parameter controlling the word distribution for each topic, and $M_i$ and $\Sigma_i$ are respectively the mean and covariance function of the Gaussian process from which $\mu_i$ are drawn.

Given a set of $N$ (not necessarily unique) members of the indexing set, $\{x_1, x_2, \dots, x_N\}$, latent probabilities for $K$ topics, as well as topics and words, are given by:
    \begin{align} 
    \mu_j &\sim \mathcal{N}\left(M_i, \Sigma_i\right), \qquad j \in \left[1..K\right] \nonumber\\
    \Phi_{z} &\sim \text{Dirichlet}\left(\beta\right), \qquad z \in \{z_1, \dots, z_K\} \nonumber\\
    z_i &\sim f(\mu_1(\vb{x}_i), \dots, \mu_K(\vb{x}_i)) , \qquad i \in \left[1..N\right] \nonumber\\
    w_i &\sim \Phi_{z_i}, \qquad i \in \qty[1..N]\label{gdrf_model_eqs}
    \end{align}
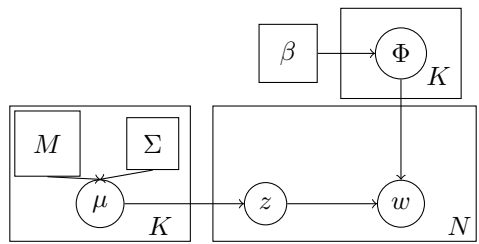
\begin{figure}[b]
    \centering
    \begin{tikzpicture}

\node (mu) at (0,0) [circle, draw] {$\mu$};
\node (z) at (2.2,0) [circle, draw] {$z$};
\node (w) at (4,0) [circle, draw] {$w$};
\node (mean) at (-.7,.8) [regular polygon,regular polygon sides=4, draw] {$M$};
\node (covariance) at (.7,.8) [regular polygon,regular polygon sides=4, draw] {$\Sigma$};

\node (psis) at (4,2) [circle, draw] {$\Phi$};
\node (b) at (2.5,2) [regular polygon,regular polygon sides=4, draw] {$\beta$};

\draw[->] (b.east) -- (psis.west);

\draw[->] (mean.south) -- (mu.north);
\draw[->] (covariance.south) -- (mu.north);

\draw[->] (mu.east) -- (z.west);
\draw[->] (z.east) -- (w.west);
\draw[->] (psis.south) -- (w.north);

\draw (-1.2,-.5) rectangle (1.2,1.3);
\draw (1.5,-.5) rectangle (5,1.3);
\draw (3.2,1.4) rectangle (4.8,2.6);

\node at (.8,-.3) {$K$};
\node at (4.78,-.3) {$N$};
\node at (4.5,1.7) {$K$};
\end{tikzpicture}
    \caption{The graphical model for GDRFs}
    \label{fig:graphical_GDRF}
\end{figure}
The graphical model for GDRFs is given in Fig. \ref{fig:graphical_GDRF}.
\section{Approximate Inference of Gaussian-Dirichlet Random Fields}
Assume we have collected a set of $N$ categorical observations $\{w_i\}$ associated with $N$ locations in the world $\{x_i\}$. We can decompose the learning of a GDRF into two steps: learning the word-topic model, and learning the latent log topic probabilities. We learn the word-topic model via Gibbs sampling. This model is similar to ROST in many ways, which is itself a spatiotemporal version of LDA. In ROST, the spatio-temporal world is discretized into cells, and the prior distribution of topics in a cell is defined by the distribution of topics in the Von Neumann neighborhood of the cell. Every word in a cell has the same prior topic distribution, independent of its exact location within the cell.

For GDRFs, the Gaussian Processes underlying the model allow us to consider topic densities, as opposed to counts, in $P(z|x)$. Normalizing $n_{-i, j}^{G(d_i)}$, the number of times topic $j$ is observed in the neighborhood of cell $d_i$, by the hypervolume $V(G(d_i)$ of cell $d_i$, we get an approximation for the mean topic density in the neighborhood of $d_i$:
\begin{equation}
    P(z_i=j|x) = \frac{\frac{n_{-i, j}^{G(d_i)}}{V(G(d_i)} + \frac{\alpha}{V(G(d_i)}}{\frac{n_{-i, \cdot}^{G(d_i)}}{V(G(d_i)} + \frac{K\alpha}{V(G(d_i)}}
\label{rost_gibbs_densities}
\end{equation}
Finally, since the Dirichlet concentration parameter $\alpha$ can also be viewed as  a smoothing ``pseudocount'', we can factor it into a scale times the hypervolume of the neighborhood, $\alpha \to \alpha V(G(d_i))$. Then, in the limit as $V(G(d_i))$ approaches zero, we get our Gibbs sampling distribution for GDRFs:

\begin{equation}
	P(z_i = j | \vb{z}_{-i}, \vb{w}, x_i) \propto \frac{n_{-i,j}^{w_i}+\beta}{n_{-i,j}^{\cdot}+W\beta} \frac{\rho_j\qty(x_i)+\alpha}{\rho\qty(x_i)+K\alpha}.
\label{gdrf_gibbs}
\end{equation}
Here $\rho_j\qty(x)$ represent the density of topic $j$ at location $x$, while $\rho(x)$ represents the observation density at location $x$. In GDRF, $\alpha$ is a pseudo-density, with the same smoothing properties as the LDA and ROST models. 

After sampling a topic for each observation, we have collected a set of $N$ categorical topics $\{z_i\}$,  associated with $N$ locations in the world $\{x_i\}$. We can use these to do approximate variational inference on the Gaussian processes.
In our generative model, we let $P(z_i|x) = \exp(\mu_i)/\sum_j\exp(\mu_j)$. 
$P(z_i | x)$ represents the topic probability at location $x$. We can use the new topics from the Gibbs sampling $\{z_i\}$ to calculate approximate $\rho_j(x_i)$ by discretizing the world. Substituting in the expression from the Gibbs sampling distribution,
\begin{equation}
	\frac{\rho_j\qty(x_i)+\alpha}{\rho\qty(x_i)+T\alpha} = \exp(\mu_j(x_i))/\sum_k\exp(\mu_k(x_i))
\label{variational_target}
\end{equation}
	or 
\begin{equation}
    \log(\rho_j\qty(x_i)+\alpha) = \mu_j(x_i) + C.
\label{variational_target_simplified}
\end{equation}
Because the softmax transformation is shift invariant, we can take $C$ to be zero. Our training inputs are the locations of the observations $\{x_i\}$, while our training targets for each Gaussian process are $\{\log(\rho_j(x_i)+\alpha)\}$. We aim to maximize the sum of the evidence lower bound (ELBO) for each GP using stochastic variational inference \cite{Hoffman2015StructuredInference}. SVI implementation was drastically simplified by using GPytorch \cite{Gardner2018GPyTorch:Acceleration}, a GP library built on top of the Python library Pytorch \cite{Paszke2017AutomaticPyTorch}, which offers simplified interfaces for automatic differentation and GPU acceleration. After a training step of the GPs, we can calculate $\rho_j$ using the same equation for the training targets. The training procedure for GDRFs is given in algorithm \ref{alg:gdrf_simple_inference}.

\begin{algorithm}[h]
\begin{algorithmic}[1]
\WHILE{true}
\FOR {$i = 1$ to $N$}
\STATE $z_i \sim P(z_i = j | \vb{z}, \vb{w})$
\ENDFOR
\STATE Update $\Phi$ according to $\{(w_i, z_i)\}$
\STATE Update $\rho_j$ according to $\{(x_i, z_i)\}$
\FOR {$j = 1$ to $K$}
\STATE $Y_j = \log\qty(\rho_j\qty(\vb{x})+\alpha)$
\STATE $\mathcal{L}_j\qty(X, Y) = \text{ELBO}\qty[\mathcal{G}_j(X), Y_j]$
\STATE Update $M_j$ and $\Sigma_j$ according to $\grad\mathcal{L}_j$
\STATE $\rho_j(x) = f_j\qty(G_1\qty(x), \dots, G_K\qty(x))$
\ENDFOR
\ENDWHILE
\end{algorithmic}
\caption{\strut GDRF Inference}
\label{alg:gdrf_simple_inference}
\end{algorithm}

In practice, the two learned components of the GDRF model (the word-topic distributions $\Phi$ and the GPs) converge at different speeds. Empirical training accuracy and time to convergence improved when allowing the Gibbs sampler and the variational inference steps to run multiple times 
before proceeding. Theoretically, $N_i$ should be chosen so each half of the algorithm converges at an equal rate during each global iteration. The Gibbs sampler is hyperparameter-free, while SVI has a single additional hyperparameter: the learning rate $\lambda$. Therefore, if $T_1$ is the mixing time of the Gibbs sampler, and $T_2/\lambda$ is the number of SVI iterations until convergence of the GPs, 
 ${N_1/ T_1} = {\lambda N_2 / T_2}$.
We found $N_1 = 50, N_2 = 5, \lambda=0.25$ allowed each half of the model sufficient time to learn from the other half, without allowing either half to converge to an undesired local optimum.


\section{Results}

\subsection{Simulated Data}

\begin{figure}[h]
\centering
\subfloat[Max Likelihood Word]{\centering\includegraphics[width=0.23\textwidth]{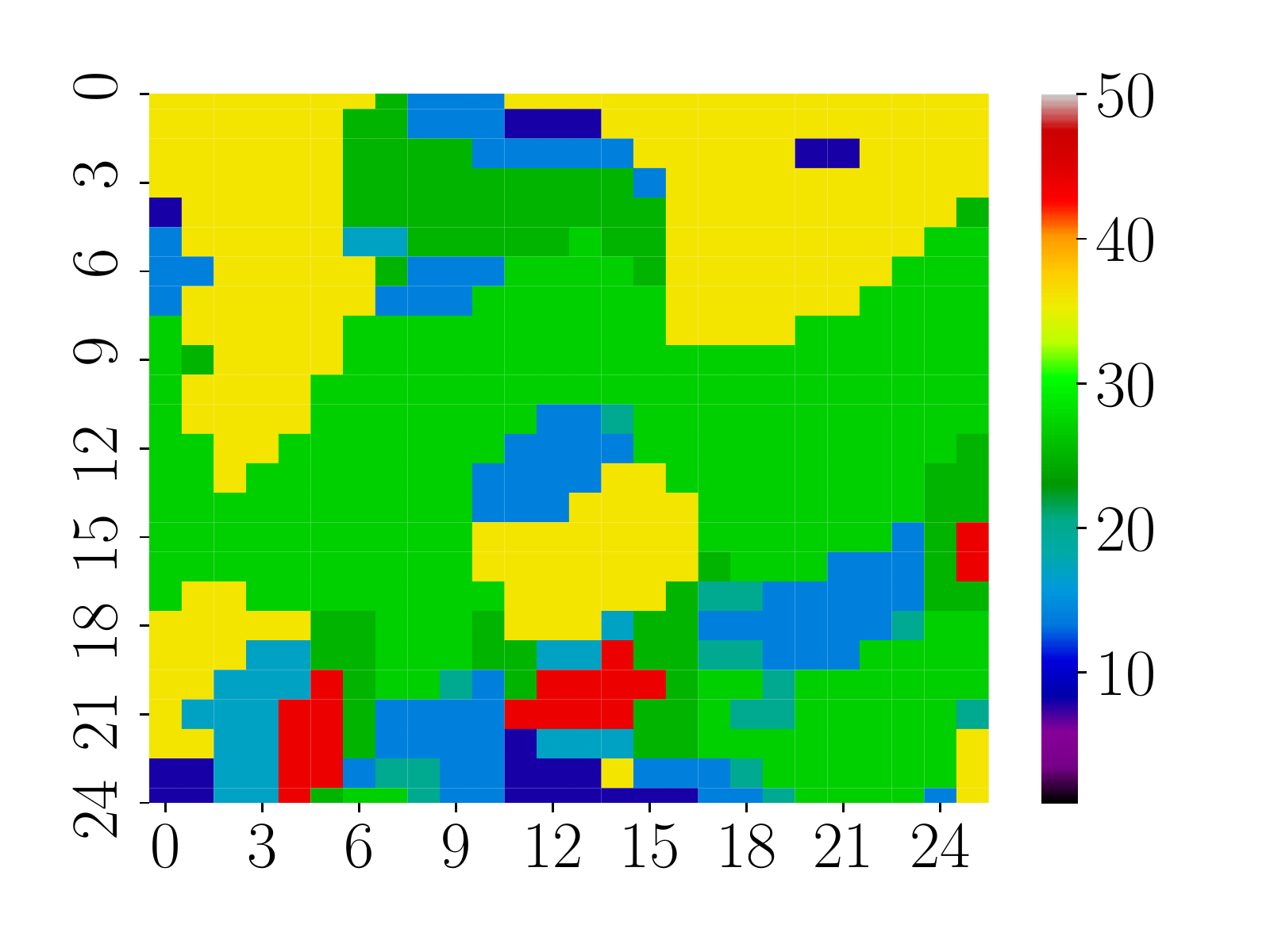}
\label{fig:sim_ml_word}}
\subfloat[Max Likelihood Topic]{\centering\includegraphics[width=0.23\textwidth]{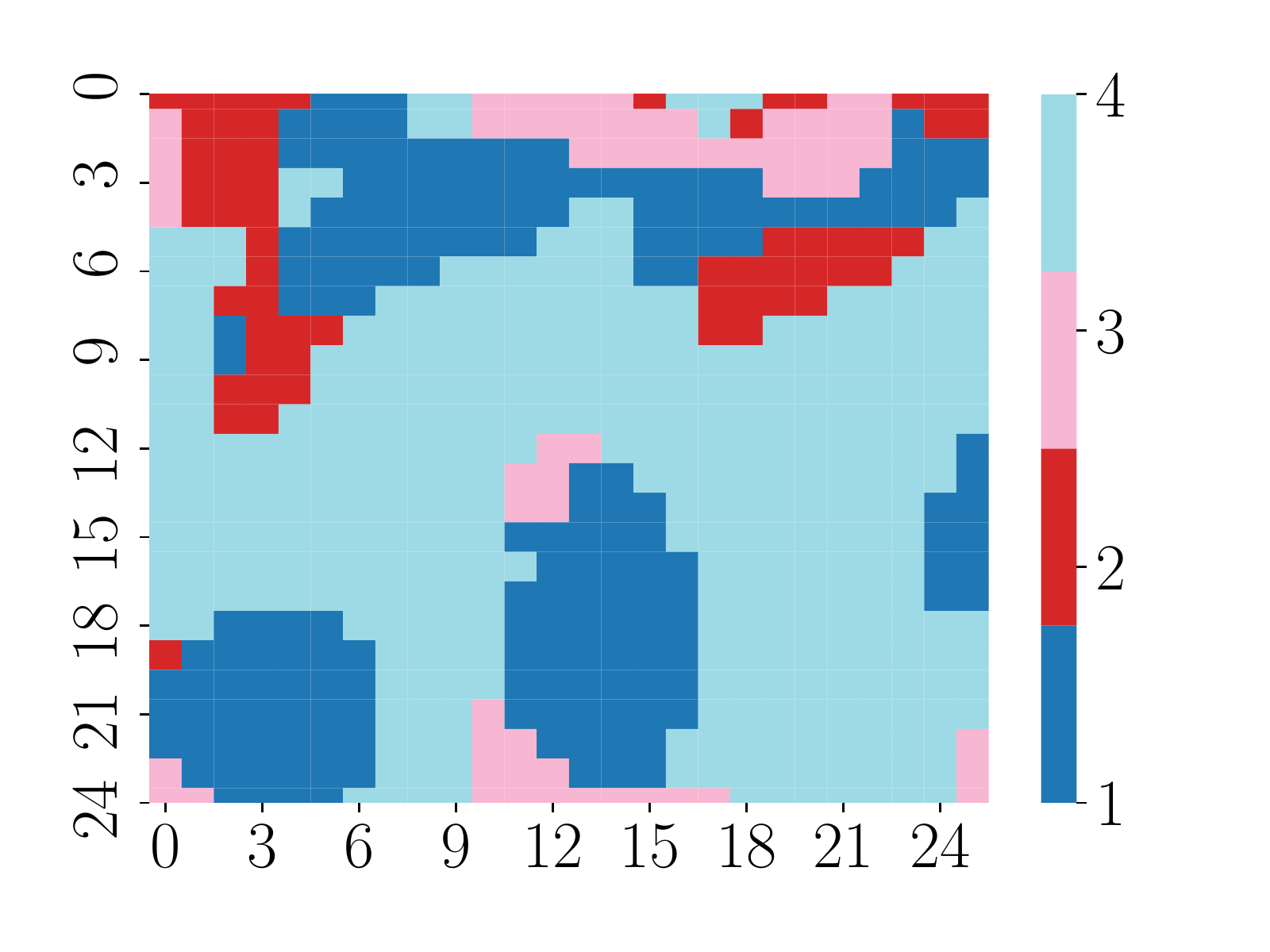}
\label{fig:sim_ml_topic}}
\caption{Forward simulation of the GDRF generative model as described in Eq. \eqref{gdrf_model_eqs} and Fig.~\ref{fig:graphical_GDRF}. The world used here is a $26 \times 26$ lattice from $0$ to $25$ in $X$ and $Y$. The means of each GP are zero, and the GPs use a Mat\'ern kernel \cite{Rasmussen:2005:GPM:1162254} with a length scale of $2.5$ in each dimension, scaled by $5$. The Dirichlet parameter $\beta$ is $0.1$}
\label{fig:sim}
\end{figure}
\begin{figure}
	\centering
	\includegraphics[width=0.9\columnwidth]{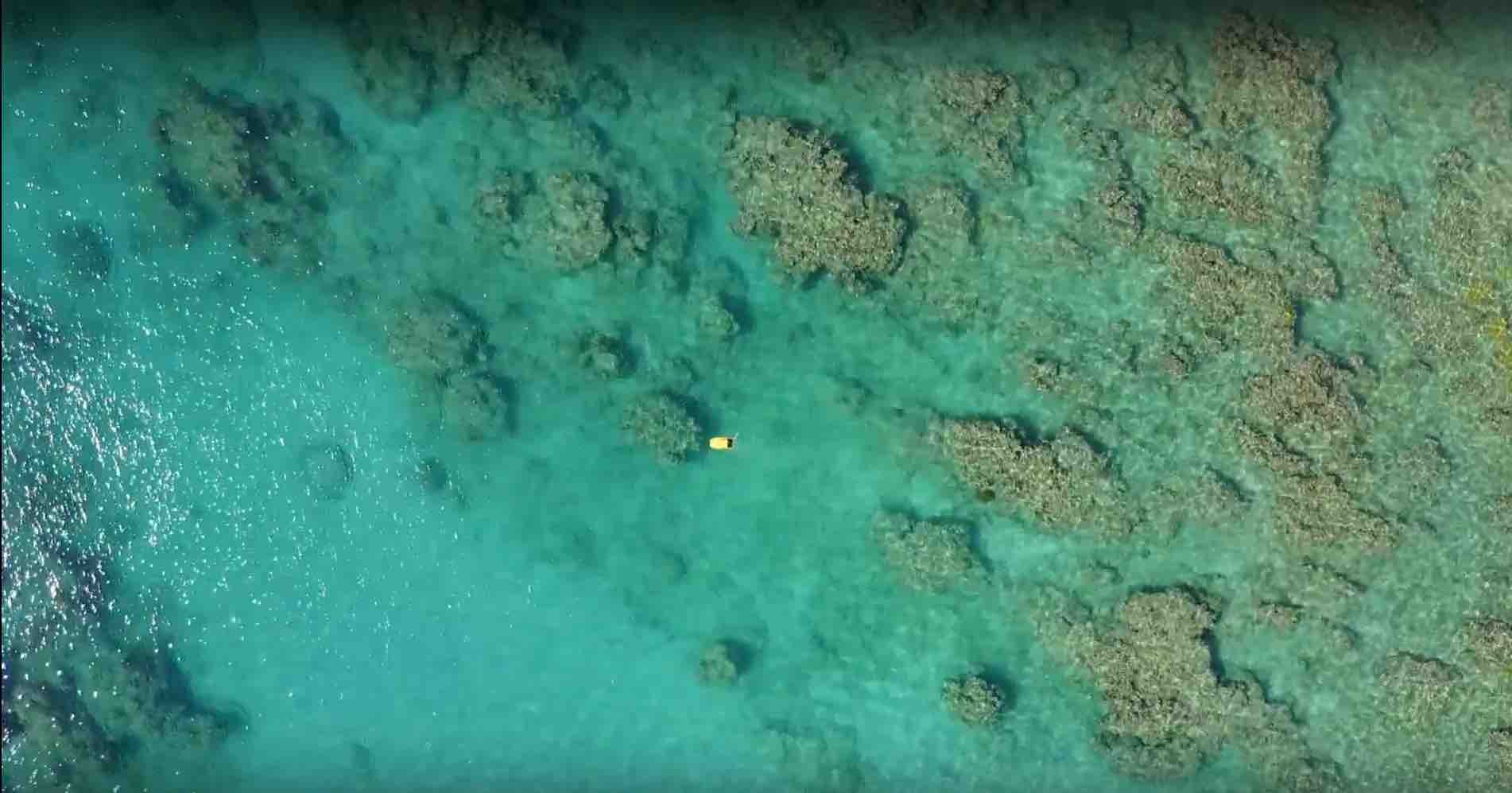}
	\caption{An example of a visually and spatially heterogeneous underwater scene. Image captured at Bellairs Research Institute, Barbados, January 2019.}
	\label{fig:corals}
\end{figure}

In order to demonstrate GDRF's ability to learn the latent topic structure underlying a set of observations, we tested both GDRF inference and ROST on a simulated dataset, shown in Fig. \ref{fig:sim}. We drew $\num[group-separator={,}]{10000}$ random observations on a $26\times26$ grid from a GDRF generative model with $4$ topics, a vocabulary of size $50$, a $\beta$ of $0.1$, zero GP mean, and a Mat\'ern kernel \cite{Rasmussen:2005:GPM:1162254} of length scale $\ell=2.5$ and overall scale $\sigma=5$. These simulation parameters were chosen because they produced  maximum likelihood word distributions, as shown in Fig. \ref{fig:sim_ml_word}, which contain realistic spatial heterogeneity. For comparison, see fig. \ref{fig:corals}, which contains coral reef structures. 

\begin{figure}[h]
\centering
\subfloat[GDRF ML Word]{\centering\includegraphics[width=0.23\textwidth]{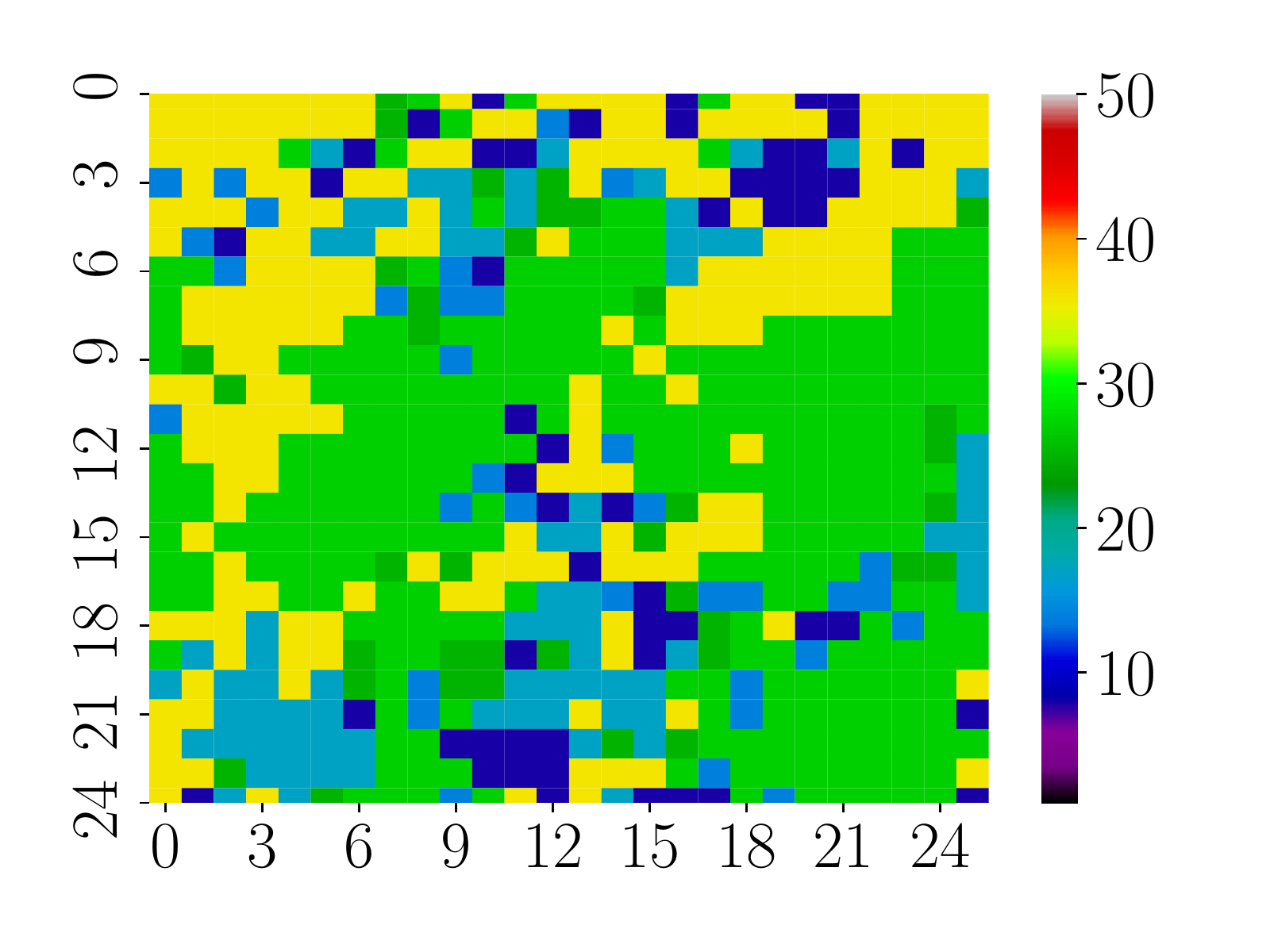}
\label{fig:gdrf_sim_ml_word}}
\subfloat[GDRF ML Topic]{\centering\includegraphics[width=0.23\textwidth]{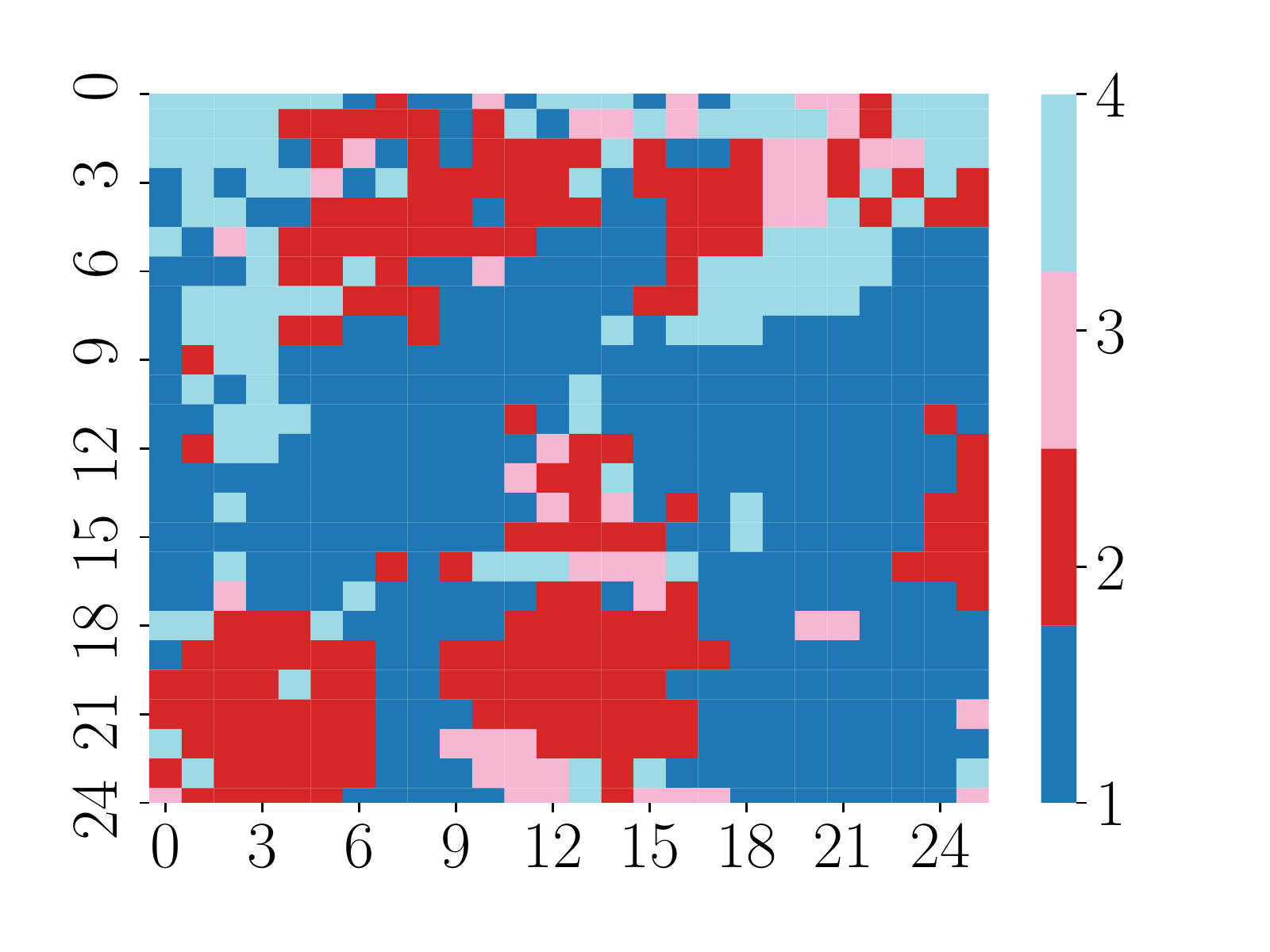}
\label{fig:gdrf_sim_ml_topic}}\newline
\subfloat[ROST ML Word]{\centering\includegraphics[width=0.23\textwidth]{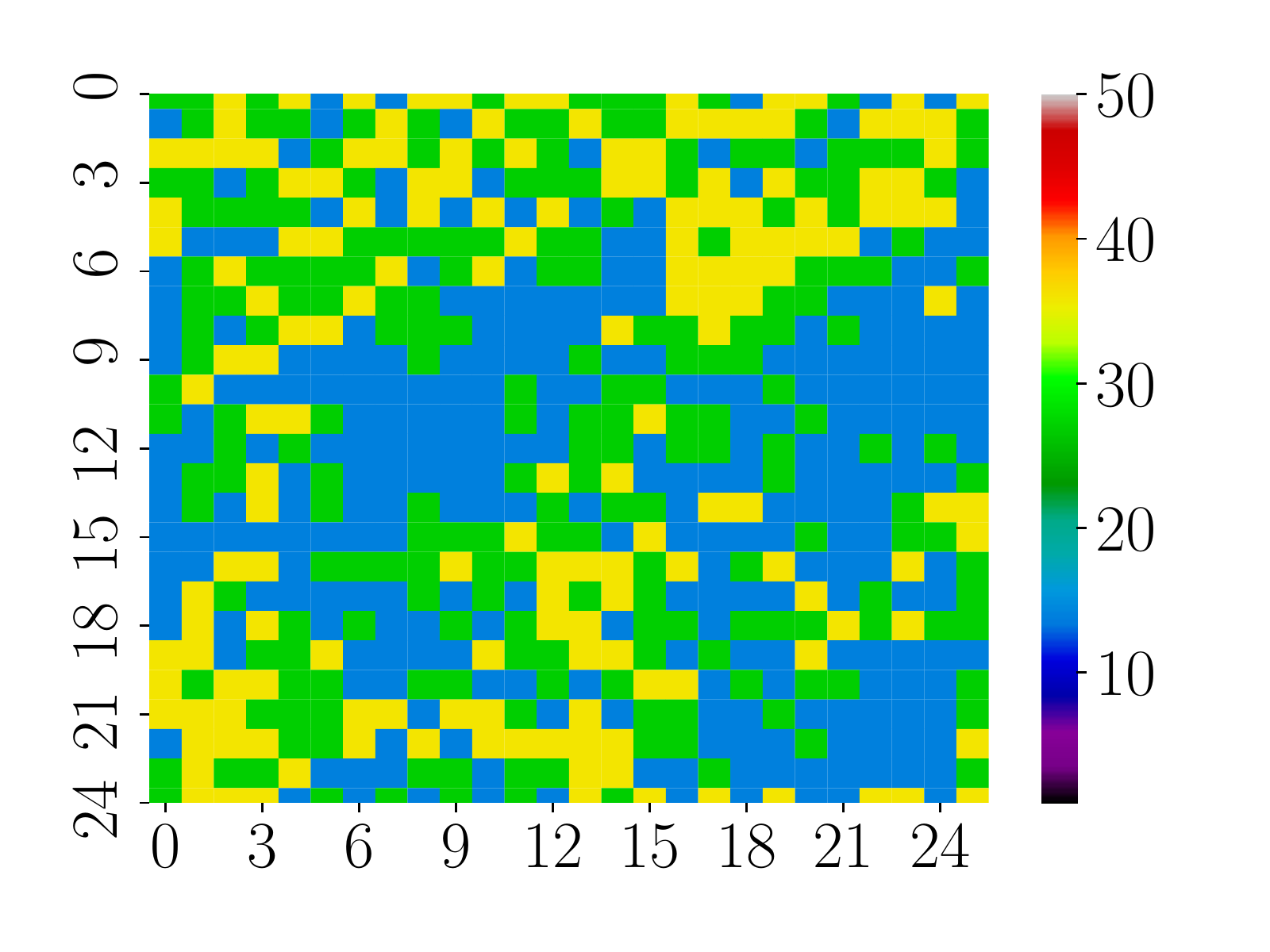}
\label{fig:rost_sim_ml_word}}
\subfloat[ROST ML Topic]{\centering\includegraphics[width=0.23\textwidth]{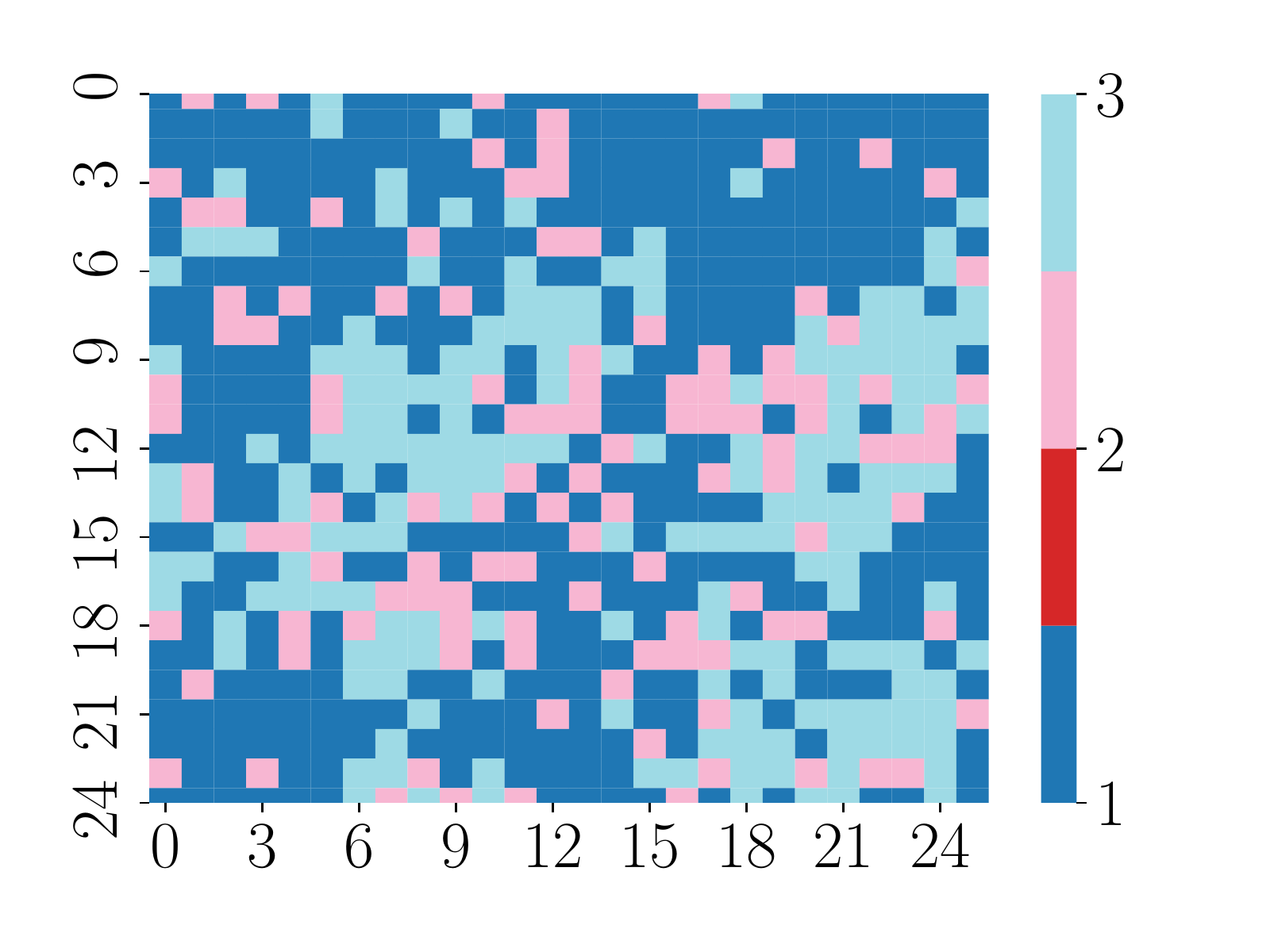}
\label{fig:rost_sim_ml_topic}}
\caption{GDRF and ROST inference on the data from Fig. \ref{fig:sim}. }
\label{fig:sim_infer}
\end{figure}

Observations drawn from the simulation shown in Fig. \ref{fig:sim} were used as inputs for both the GDRF model and the ROST model. The maximum likelihood topics for both models (shown in Fig. \ref{fig:gdrf_sim_ml_topic}) are visually similar to the ground truth maximum likelihood topics in Fig. \ref{fig:sim}. But the GDRF's inferred maxmimum likelihood topics in Fig. \ref{fig:gdrf_sim_ml_topic} have a much stronger resemblance to the ground truth in Fig. \ref{fig:gdrf_sim_ml_topic} than do ROST's in Fig. \ref{fig:rost_sim_ml_topic} (modulo the swapping of topic labels and colors). In addition, ROST learns a factorization in which three of the four topics are always most likely, while GDRF's factorization correctly models the observations with four topics.  The GDRF model also captures small scale variation (on the order of a couple of spatial cells) in the latent topic field that ROST fails to pick up. Finally, note that the inputs to both of these models are the observations, as in Fig. \ref{fig:sim_ml_word}, but the similarity between the ground-truth maximum likelihood topics in Fig. \ref{fig:sim_ml_topic} and the GDRF's inferred maximum likelihood topics in Fig. \ref{fig:gdrf_sim_ml_topic} is the product of a completely unsupervised training process.

\begin{figure}[t]
    \centering
    \subfloat[Word AFMI]{\centering\includegraphics[width=0.23\textwidth]{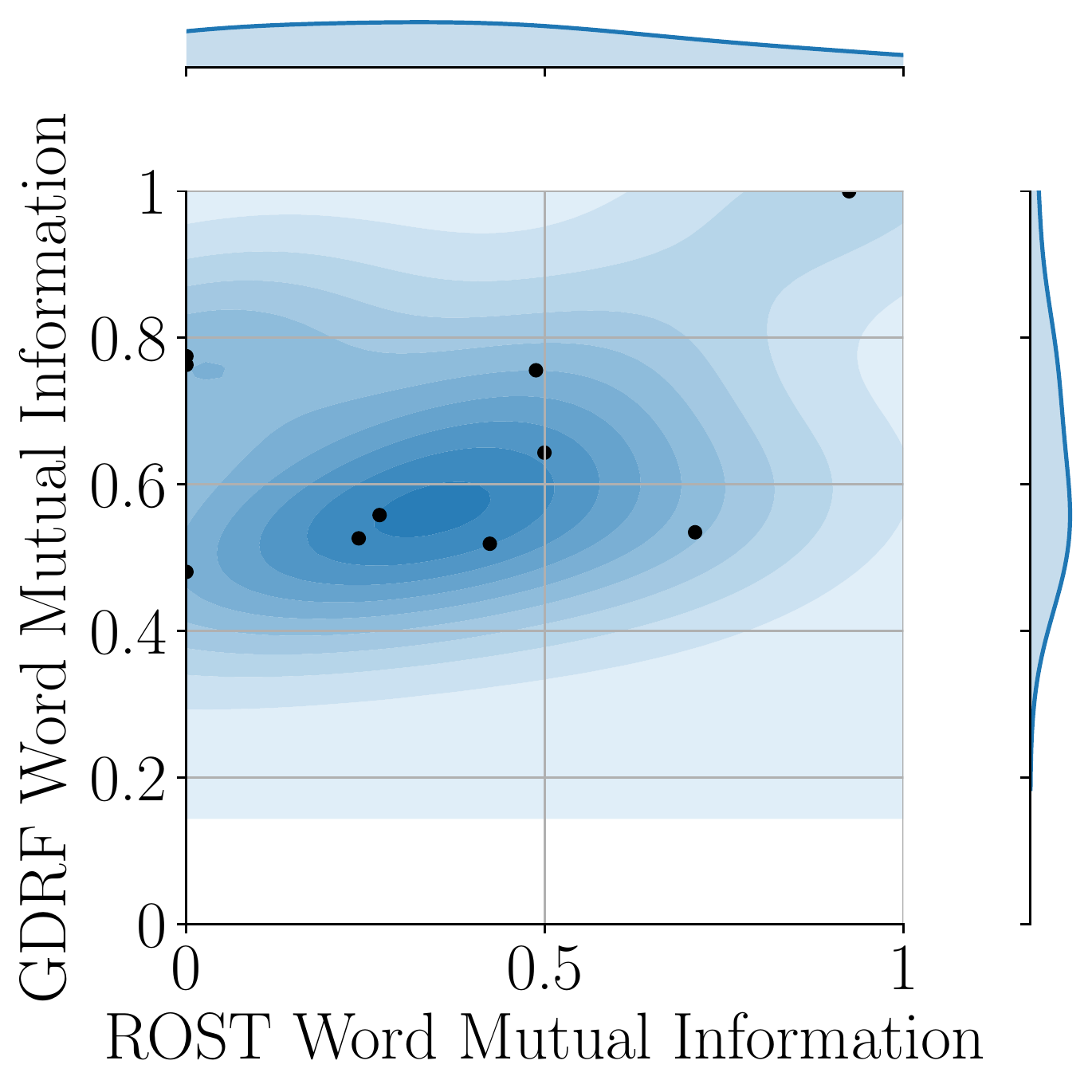}\label{fig:word_mmi}}
    \subfloat[Topic AFMI]{\centering\includegraphics[width=0.23\textwidth]{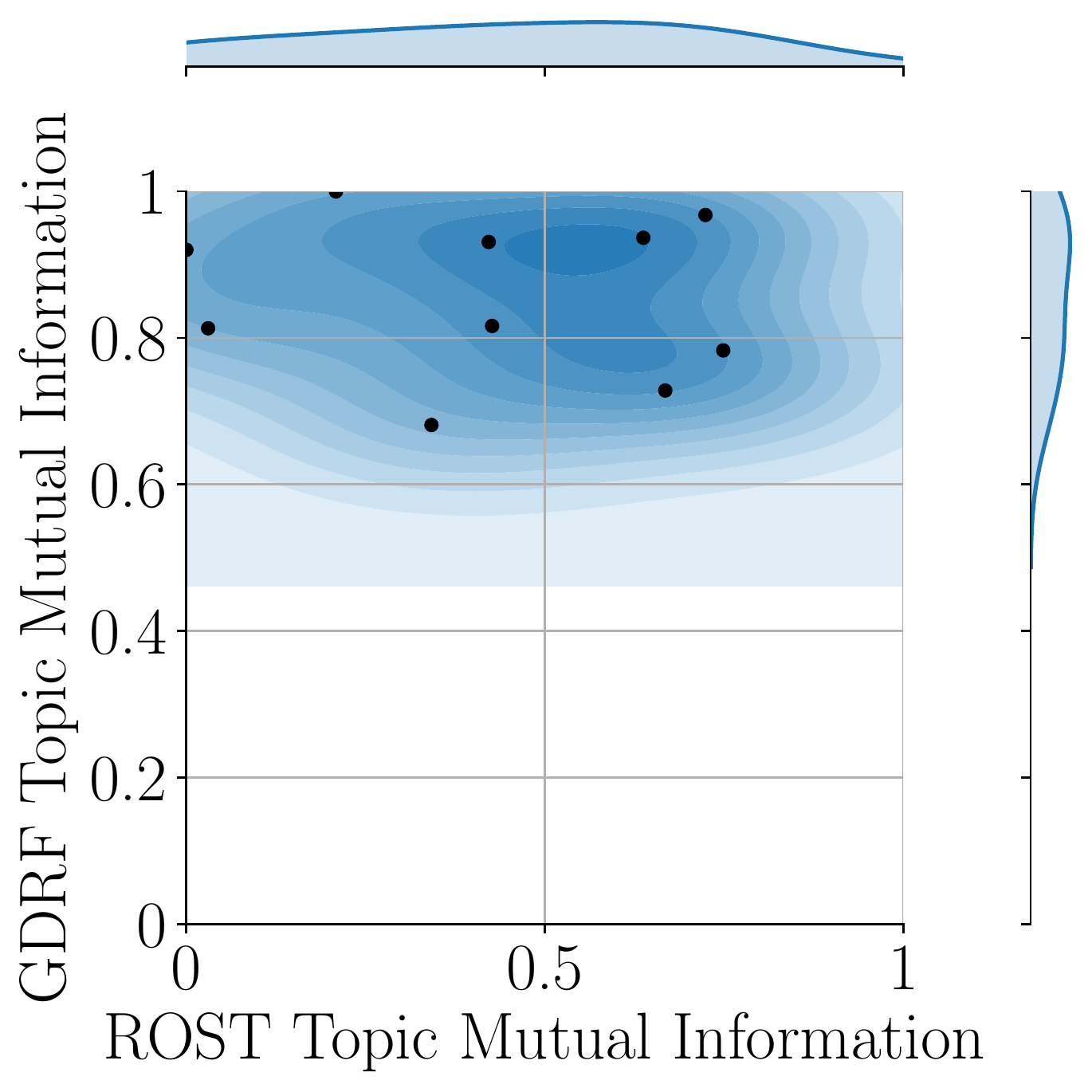}\label{fig:topic_mmi}}
    \caption{Approximate Fractional Mutual Information (AFMI) for GDRF and ROST on 10 random data sets. \ref{fig:topic_mmi}: Topic AFMI. \ref{fig:word_mmi}: Word AFMI. ROST AFMI is on the X-axis, while GDRF AFMI is on the Y-axis.}
    \label{fig:mmi}
\end{figure}
To numerically evaluate the ability of these two models to capture complex spatial heterogeneity, we generated $10$ simulation of $\num[group-separator={,}]{10000}$ observations on an $11\times 11$ grid, from GDRFs with $3$ topics, a vocabulary of size $15$, $\beta = 0.1$, $M = 0$, and Mat\'ern kernels with $\sigma = 5$ and $\ell = 12.5$. We trained both GDRFs and ROST on each simulation, and we computed an approximate fractional mutual information (AFMI) score between the inferred and ground truth maximum likelihood topics and words for both models:
\begin{align}
\text{AFMI}(M_{\text{model}}, M_{\text{gt}}) &= \frac{I(M_{\text{model}}, M_{\text{gt}})}{I(M_{\text{gt}}, M_{\text{gt}})} \\
I(X, Y) &= \sum_x\sum_yP(x, y)\log\qty(\frac{P(x, y)}{P(x)P(y)})
\end{align}
Here the marginal probabilities $P(M = i)$ are  proportional to the number of cells in the maximum likelihood map $M$ with category $i$, and the joint probabilities $P(M_{a} = i, M_{b} = j)$ are proportional to the number of cells with topic $i$ in map $a$ and topic $j$ in map $b$. Note that AFMI scores range from zero to one, with zero representing two maps with no mutual information and one representing two identical maps. 

The resultant AFMI scores are shown in Fig. \ref{fig:mmi}. GDRF has consistently high AFMI scores for its topic maps in Fig. \ref{fig:topic_mmi}, and slightly lower AFMI scores for its word maps Fig. \ref{fig:word_mmi}. In contrast, ROST has AFMI scores that vary along the entire scale for both topics and words. In both cases, GDRFs almost always get higher AFMI scores than ROST. The GDRF model is capable of learning complex, heterogeneous latent structures from densely sampled 2D categorical fields, and it learns those structures better than ROST.

%

\subsection{Phytoplankton Taxa Dataset}
\begin{figure}[t]
    \centering
    \subfloat[Observed Taxon Distributions]{\centering\includegraphics[width=0.23\textwidth]{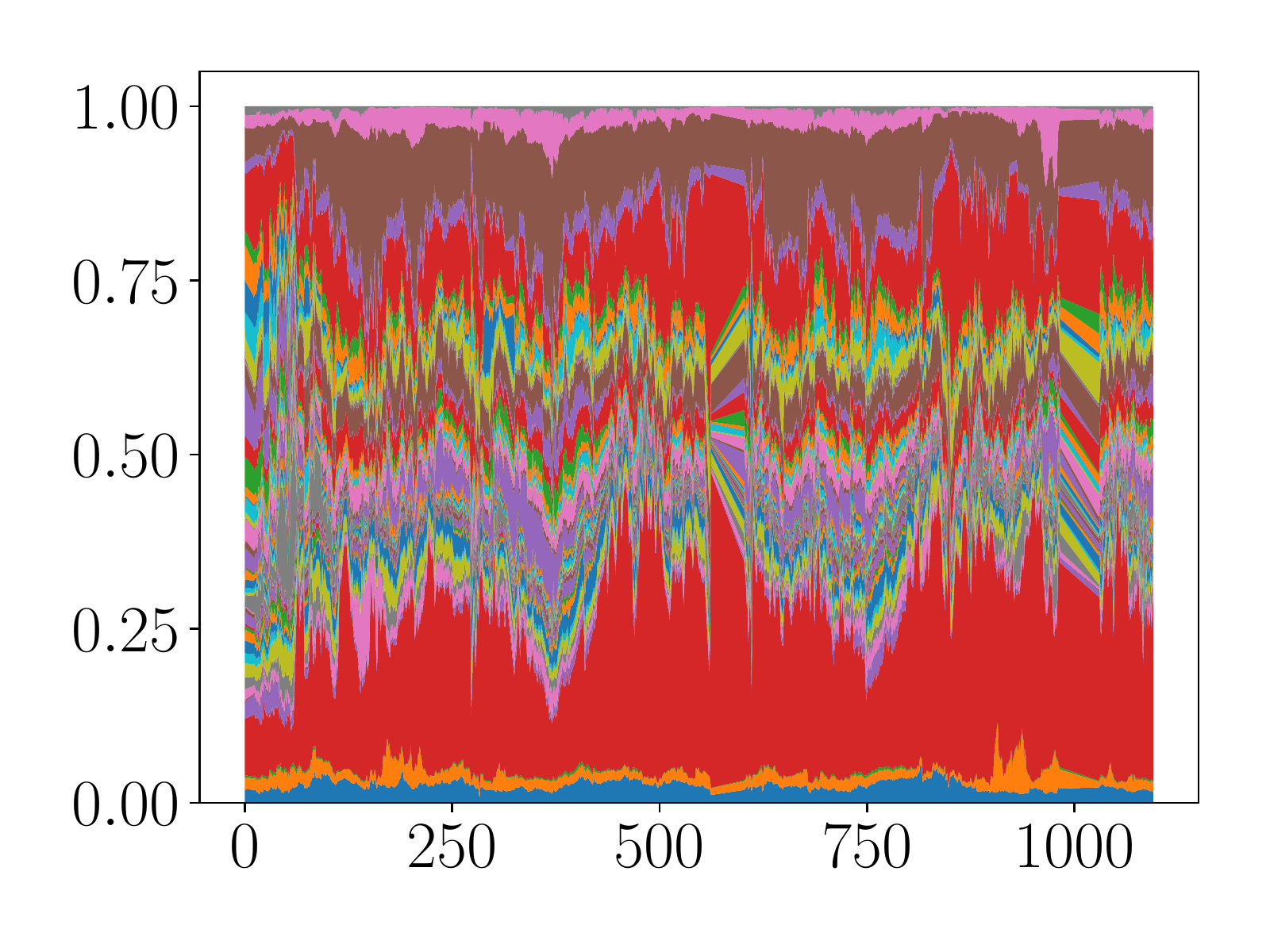}\label{fig:mvco_obs_word}}
    \subfloat[KL Divergence]{\centering
    \includegraphics[width=0.23\textwidth]{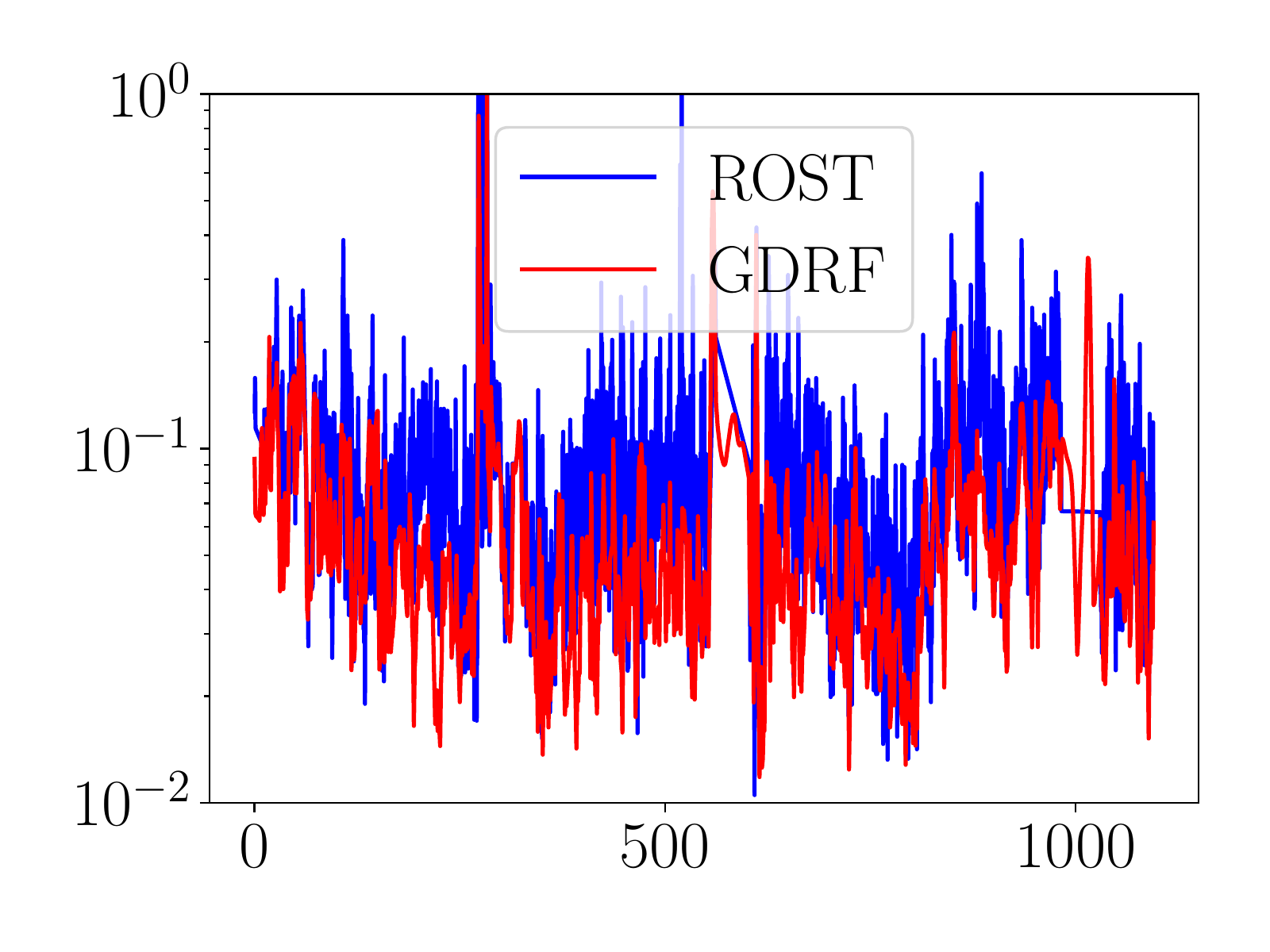}\label{fig:mvco_kldiv}}\newline
	\subfloat[ROST Taxon Distributions]{\centering\includegraphics[width=0.23\textwidth]{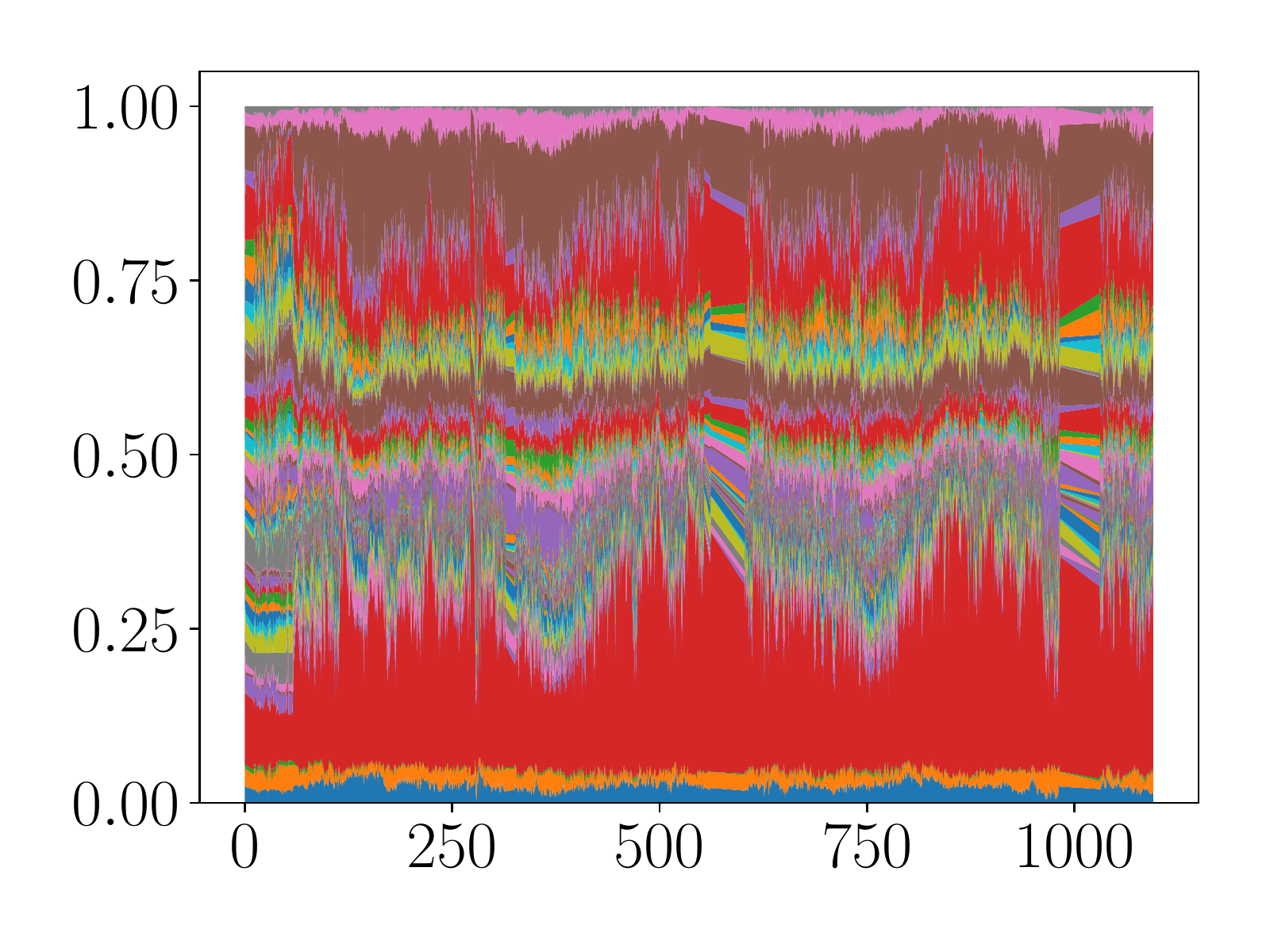}\label{fig:mvco_rost_word}}
	\subfloat[ROST Topic Distributions]{\centering\includegraphics[width=0.23\textwidth]{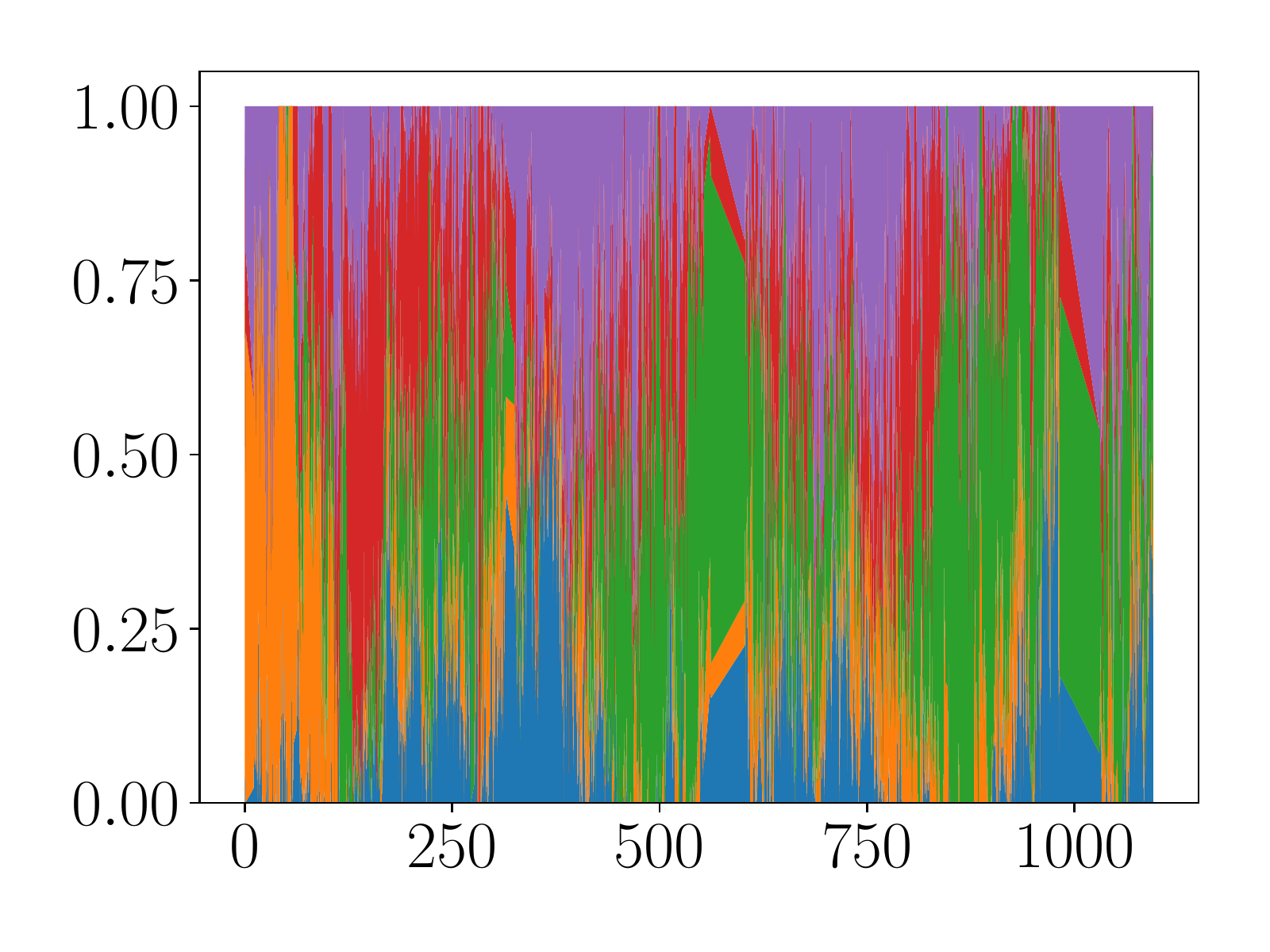}\label{fig:mvco_rost_topic}} \newline
    \subfloat[GDRF Taxon Distributions]{\centering\includegraphics[width=0.23\textwidth]{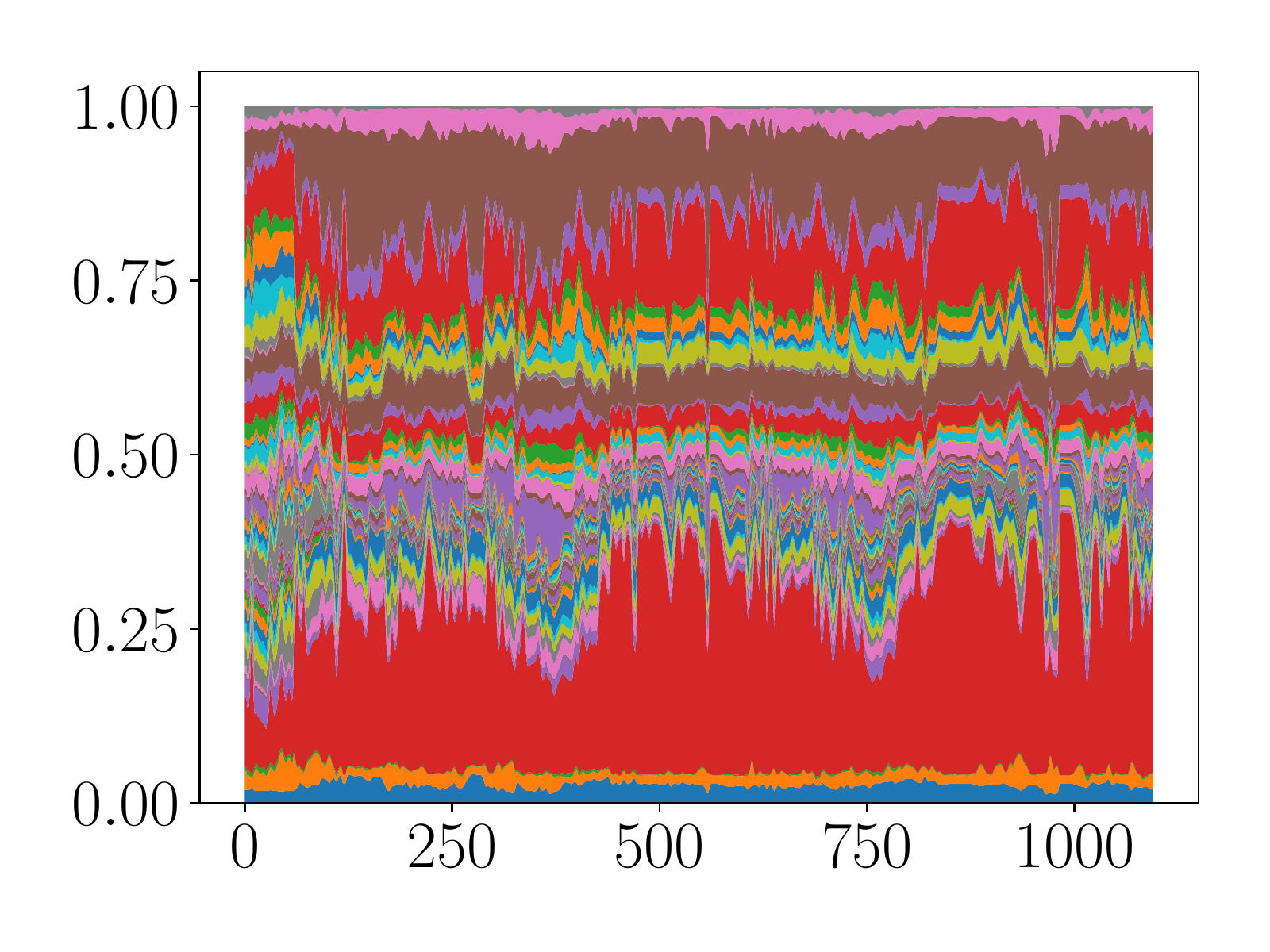}\label{fig:mvco_gdrf_word}}
    \subfloat[GDRF Topic Distributions]{\centering\includegraphics[width=0.23\textwidth]{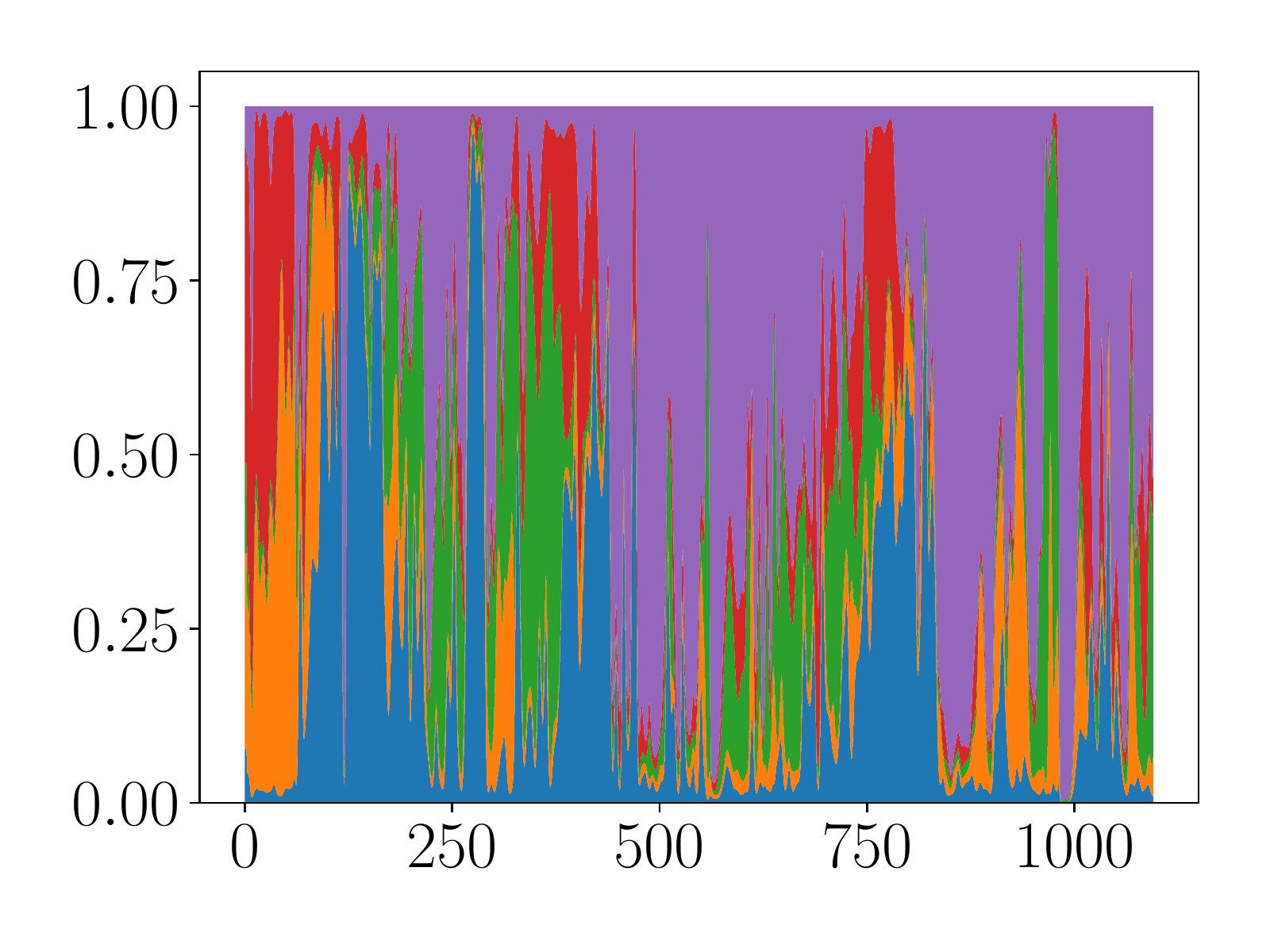}\label{fig:mvco_gdrf_topic}}
    \caption{MVCO phytoplankton taxa dataset. Ground truth taxon observations are produced by running phytoplankton images through a random forest classifier. The output labels are used as training data for both models. \ref{fig:mvco_obs_word}: Observed phytoplankton taxon distributions. \ref{fig:mvco_kldiv}: Mean KL Divergence between ROST (blue) or GDRF (red) inferred taxon distribution and ground-truth observed taxon distributions. \ref{fig:mvco_rost_word}: Inferred taxon distributions for ROST model. \ref{fig:mvco_rost_topic}: Inferred latent topic distributions for ROST model. \ref{fig:mvco_gdrf_word}: Inferred taxon distributions from GDRF model. \ref{fig:mvco_gdrf_topic}: Inferred latent topic distributions for GDRF model. The X-axis represents time in days. The Y-axis represents fraction of total probability. }
    \label{fig:mvco}
\end{figure}

The Martha's Vineyard Coastal Observatory (MVCO) is a research station off the coast of Martha's Vineyard, Massachusetts, providing long time series of oceanographic and meteorological measurements \cite{Austin2002AObservatory}. In addition to typical oceanographic sensors, the MVCO is equipped with an Imaging FlowCytobot\cite{Olson2007AFlowCytobot}, which takes high-throughput images of phytoplankton. These  images are both selectively expert-annotated, and automatically labelled using a machine learning-based classifier \cite{Sosik2007}. 

We used a random subset of $\num[group-separator={,}]{100000}$ images from three years of MVCO phytoplankton data, from $2013$ to $2016$, labeled by taxon using a random forest classifier, as training inputs for GDRF and ROST. The results are shown in Fig. \ref{fig:mvco}. Both GDRFs and ROST generate inferred word distributions (Figs. \ref{fig:mvco_rost_word} and \ref{fig:mvco_gdrf_word} ) that are visually similar to the observed distribution of phytoplankton taxa (Fig. \ref{fig:mvco_obs_word}). But Fig \ref{fig:mvco_kldiv} shows GDRF produces a distribution with a lower mean KL-divergence from the observed distribution, $0.064$, than ROST, $0.095$. This shows GDRF learns a more accurate model for the observations than ROST.

\begin{figure}[h]
    \centering
	\includegraphics[width=0.38\textwidth]{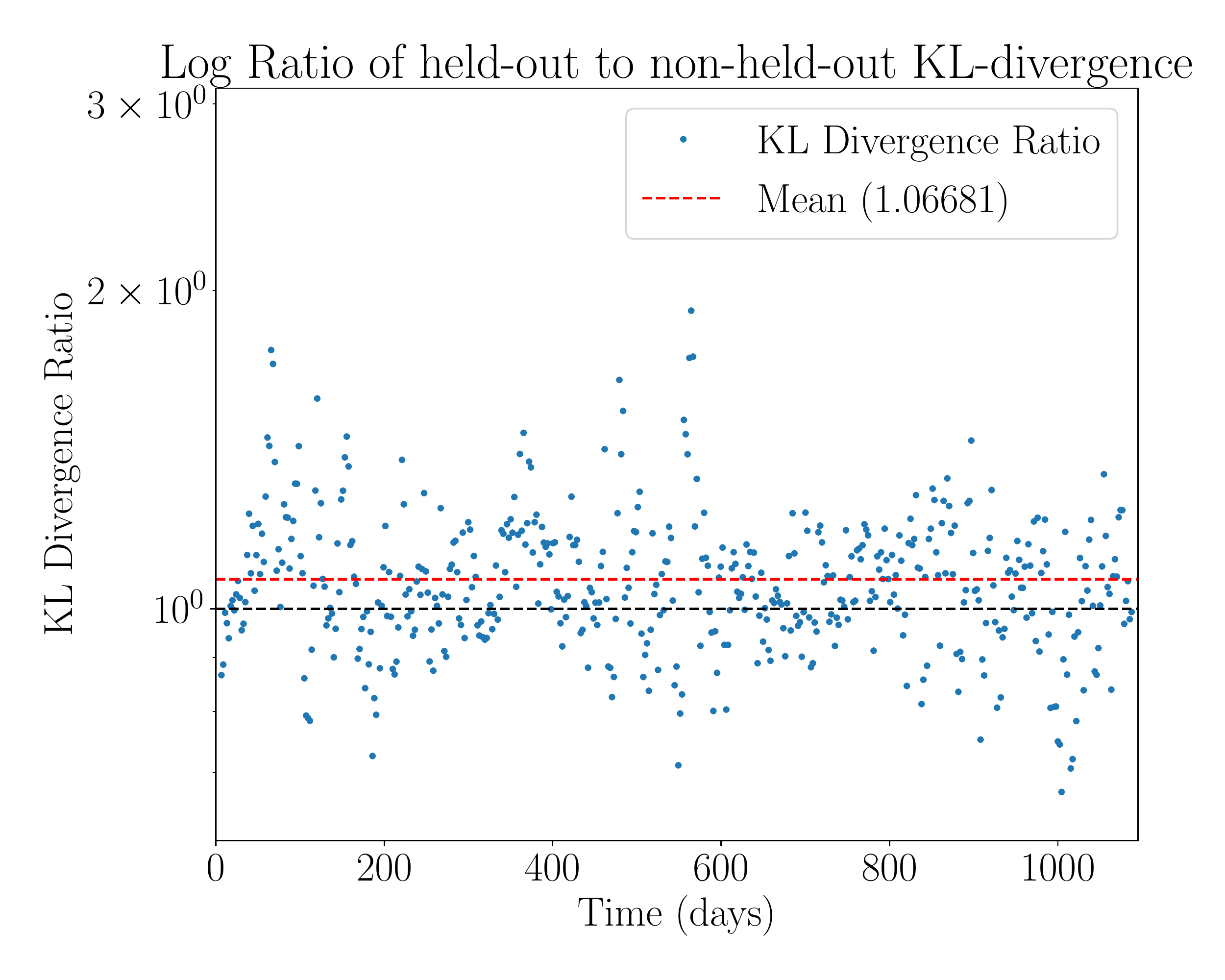}
	\caption{Comparison of GDRF accuracy with and without held-out data. For each data point, a roughly 10 day window of MVCO data was held out of the training data set, to produce the held-out model. The average KL divergence between the model and the ground-truth in the window was calculated. The y-axis represents the ratio between this held-out KL divergence from ground truth and the non-held-out model's average KL divergence from ground truth within the same window. Values above one imply that the non-held-out model has a smaller KL divergence from ground truth than the held-out model.} 
	\label{fig:h_mvco}
\end{figure}

We demonstrate GDRF's ability to extrapolate by comparing the model's performance after training with and without withheld subsets of the data. The three years of phytoplankton taxon data are divided into $500$ equal-time chunks. A held-out window five chunks wide (approximately $10$ days) is slid across the dataset, and the model is trained without that chunk of data. The mean KL divergence between the model and the observation distribution within this window is calculated, and fig. \ref{fig:h_mvco} plots the log ratio between this window-mean divergence in the held-out and non-held-out models. On this figure, a ratio of $1$ implies than both the model trained with held-out data and the model trained without held-out data predicted taxon distributions equally as far from the observed distribution (as measured using KL divergence); a ratio above $1$ implies that the model trained with held-out data does worse. The mean ratio between the models is $1.07$. The model trained without data in a particular region predicts a taxon distribution that is less accurate than a model trained to reproduce the observed distribution. But the magnitude of the inaccuracy is relatively low, with no held-out windows exhibiting more than a factor of 2 increase in KL divergence from observations. Since the GDRF model trained without any held-out data performs well when compared to ROST, predictions from a GDRF model in unobserved regions or periods of time are likely accurate enough to use for adaptive sampling and informative path planning. 
%
%
%
%
\section{Conclusion}
The GDRF model provides both a generative process for spatially heterogeneous categorical observation fields, and a method for extrapolating from categorical observations underpinned by semantically meaningful Gaussian processes. By factoring the high-dimensional distribution over observation categories into a low-dimensional distribution over topics, GDRFs are able to take advantage of the expressiveness of Gaussian process models for categorical data. We have demonstrated that GDRFs are capable of extracting meaningful structure from 2D spatial fields and interpolating on data sets with temporal gaps. 

The effectiveness of GDRF inference on the MVCO phytoplankton taxon dataset demonstrates the viability of GDRFs for modeling biological systems in the ocean. 
We have also experimented with neural network classifiers capable of distinguishing over 140 phytoplankton taxa. Future work will explore the use of GDRFs to model 2-D and 3-D phytoplankton fields, labeled by neural network.

In this paper, we described a batch training procedure for learning GDRF models. Future work will develop techniques for doing online inference, enabling many applications such as IPP  for mapping of map categorical fields autonomously.
\bibliographystyle{IEEEtran}
\bibliography{IEEEabrv,bibliography,girdhar}
\end{document}